\RequirePackage{fix-cm}
\documentclass[smallextended]{svjour3}       
\smartqed  
\usepackage{graphicx}
\usepackage{caption}
\usepackage{subcaption}
\usepackage{amsmath}
\usepackage{booktabs,amssymb,verbatim,fancyhdr,mathrsfs}
\newcommand\cov{\mathop{\rm cov}}

\newcommand\vecX{\mathbf{x}}
\newcommand\vecY{\mathbf{y}}
\newcommand\vecK{\mathbf{k}}

\newcommand\vecT{\boldsymbol{\theta}}

\newcommand\matC{\mathbf{C}}
\newcommand\matI{\mathbf{I}}
\newcommand\matK{\mathbf{K}}
\newcommand\matL{\mathbf{L}}
\newcommand\matR{\mathbf{R}}
\newcommand\matV{\mathbf{V}}
\newcommand\matX{\mathbf{X}}

\newcommand\bbX{\mathbb{X}}
\newcommand\bbR{\mathbb{R}}

\newcommand\iD{d} 
\newcommand\sS{n} 
\journalname{AMAI}
\begin{document}

\title{Large Scale Variable Fidelity Surrogate Modeling \thanks{The research was supported by the the Russian Science Foundation grant (project 14-50-00150).}
}


\author{Zaytsev A.        \and
        Burnaev E.
}


\institute{A. Zaytsev \at
              Skolkovo Institute of Science and Technology, Moscow, Russia, 143026 \\
              Institute for Information Transmission Problems, Moscow, Russia, 127051 \\
              DATADVANCE LLC, Moscow, Russia, 117246 \\
              \email{likzet@gmail.com}           
           \and
           E. Burnaev \at
             Skolkovo Institute of Science and Technology, Moscow, Russia, 143026 \\
              Institute for Information Transmission Problems, Moscow, Russia, 127051 \\
              DATADVANCE LLC, Moscow, Russia, 117246 \\
              \email{e.burnaev@skoltech.ru}
}

\date{Received: date / Accepted: date}

\maketitle

\begin{abstract}
Engineers widely use Gaussian process regression framework to construct surrogate models aimed to replace computationally expensive physical models while exploring design space.
Thanks to Gaussian process properties we can use both samples generated by a high fidelity function (an expensive and accurate representation of a physical phenomenon) and a low fidelity function (a cheap and coarse approximation of the same physical phenomenon) while constructing a surrogate model.
However, if samples sizes are more than few thousands of points, computational costs of the Gaussian process regression become prohibitive both in case of learning and in case of prediction calculation.
We propose two approaches to circumvent this computational burden: one approach is based on the Nystr\"om approximation of sample covariance matrices and another is based on an intelligent usage of a blackbox that can evaluate a~low fidelity function on the fly at any point of a design space.
We examine performance of the proposed approaches using a number of artificial and real problems, including engineering optimization of a rotating disk shape.

\end{abstract}

\section{Introduction}
\label{introintro}

Computational modeling is widely adopted in various branches of engineering~\cite{velten2009mathematical}.
The aim of the computational modeling is to replace costly field experiments 
with evaluations of a less costly computational code.
While wide adoption of the mathematical modeling significantly reduces time and costs required to perform a design process in the industrial engineering,
in many cases it still requires several days and HPC resources to perform an experiment~\cite{forrester}.

Nowadays engineers often construct surrogate models to replace expensive computational models with cheap but sufficiently accurate approximations~\cite{forrester,gorissen2010surrogate,erofeev2016ann}.
An engineer generates a~sample of points, evaluate values of a high fidelity computational code (a high fidelity function) at these points, and then use the generated sample and machinery of regression analysis to construct a surrogate model.

Gaussian process regression is an attractive framework for construction of nonlinear regression models~\cite{forrester,rasmussen2006} with a number of guaranteed theoretical properties~\cite{burnaev2013bernstein,van2008rates}.
Constructed model can be used to speed-up evaluations~\cite{sterling2014approximation,alestra2014df,grihon2013structureOPT}, for surrogate-based optimization~\cite{forrester,martin2005use}, uncertainty quantification~\cite{kawai2013kriging}, sensitivity analysis \cite{panin1,panin2,panin3} and adaptive design of experiments~\cite{burnaev2015adaptive}. 
Maturity of this approach is further confirmed not only by numerous applications but also by availability of software packages that are dedicated to surrogate modeling and include Gaussian process regression-based approaches~\cite{belyaev2016gtapprox,gorissen2010surrogate,ageev2016minimization,bergh2013ansys}.

Nice property of Gaussian process regression is an ability to treat variable fidelity data~\cite{forrester2007multi,kennedy2000predicting,qian2006building,han2013improving,doyen1988porosity,burnaev2015surrogate}: one can construct a surrogate model of a~high fidelity function using data both from high and low fidelity sources (e.g., high fidelity evaluations can be obtained using a computational code with a~fine mesh, and low fidelity evaluations can be obtained using the same computational code with a~coarser mesh).
Recent results provide theoretical analysis of obtained models~\cite{zaytsev16arxive,zhang2015doesn} and of parameters estimates~\cite{burnaev2013bernstein}.
Gaussian process based variable fidelity modeling shares many common ideas with
multi-output Gaussian process regression~\cite{alvarez2011computationally,boyle05dependent,higdon2008computer,chang2014fast}. 

For data with no specific structure we need $O(\sS^2)$ memory to store a surrogate model and $O(\sS^3)$ operations to construct it.
Due to this computational complexity usually not more than a few thousands of points are used when training Gaussian Process regression.
In the variable fidelity scenario samples are often large, as one evaluation of a low fidelity function is usually significantly cheaper than one evaluation of a high fidelity function.

Currently there are several ways to reduce memory and computational requirements for Gaussian process regression.
Nystr\"om approximation~\cite{drineas2005nystrom} is a popular approach to perform large sample Gaussian process regression inference~\cite{quinonero2005unifying,foster2009stable,sun2015review}.
The idea is to select a subsample of a full sample for which we can still perform Gaussian process regression inference, and then approximate the full sample covariance matrix and its inverse by some combination of the covariance matrix for the selected subsample and the cross-covariance matrix between points from the selected subsample and from the full sample.
Bayesian approximate inference provides an alternative fast estimate of the full sample likelihood that is then optimized to estimate model parameters~\cite{titsias2009variational,hensman2013gaussian}. 
Another popular approach with thoroughly investigated theoretical properties is a covariance tapering~\cite{furrer2006covariance,shaby2012tapered}:
we set a covariance between points equal to zero in case a distance between them is above some threshold, so in such a way we obtain sparse covariance matrices, and we can efficiently process them with appropriate routines.
Hierarchical models also can alleviate the computational burden, as they split the sample into separate subsamples. However, in this case exact inference is possible also only if we make some specific assumptions about their covariance structure~\cite{park2010hierarchical,shi2005hierarchical,banerjee2008gaussian}. 
In addition, fast exact inference is possible if training data has some specific structure:
for example, in \cite{belyaev2014exact,burnaevKap} authors describe an exact inference scheme to construct Gaussian process regression in case of samples with a Cartesian product structure.
However, as far as we know there are no approaches to large scale variable fidelity Gaussian process regression in case of data without any specific structure.

Another issue with Gaussian process regression is its bad extrapolation properties: 
the model prediction at a~new point is a weighted sum of function values at the available training points with weights defined by covariances between these points~\cite{rasmussen2006}; 
i.e., the prediction can be determined only locally near the training points, and we need to be careful with test points that are far away from the training sample.

We propose two approaches that mitigate the sample size limitation and improve extrapolation properties of variable fidelity Gaussian process regression.
The first approach adopts the Nystr\"om approximation and relies on the results obtained for single fidelity data in the Sparse Gaussian process regression framework~\cite{foster2009stable,kumar2012sampling}.
The second approach uses a low fidelity function blackbox that provides low fidelity function evaluations on the fly: we improve prediction of a surrogate model at a new point using the low fidelity function value at this point.
While for heuristic models it is a common practice to incorporate a low fidelity function blackbox in this way~\cite{alexandrov2000first,madsen2001multifidelity,sun2010two,zahir2012variable,xu1992integrating}, Gaussian process regression doesn't support its direct usage. 
As we are able to evaluate a low fidelity function at any point of a design space, we avoid using a large sample to cover the whole design space.
Instead, we just need to obtain a low fidelity sample that is sufficient for accurate estimation of Gaussian process regression model parameters.

We investigate computational complexity and compare accuracies of the proposed approaches using real and artificial data.
The real problem at hand is an optimization of a~rotating disk in an aircraft engine.
The disk shape optimization problem remains challenging and often involves usage of surrogate modeling of maximal stress and radial displacement of the disk~\cite{huang2011optimal,farshi2004optimum}.
We compare four approaches to construct surrogate models: Gaussian process regression, Gaussian process regression for variable fidelity data, and  approaches presented in this paper --- sparse Gaussian process regression for variable fidelity data and Gaussian process regression for variable fidelity data with an available low fidelity function blackbox.

The paper contains the following sections:
\begin{itemize}
\item Section~\ref{sec:gpRegression} describes the Gaussian process regression framework;
\item Section~\ref{sec:variableFidelityGp} outlines the variable fidelity Gaussian process regression framework;
\item Section~\ref{sec:sparse} proposes an approach to construct sparse Gaussian process regression for variable fidelity data;
\item Section~\ref{sec:blackbox} describes our approach to variable fidelity Gaussian process regression with a low fidelity function blackbox;
\item Section~\ref{sec:problem} provides results of computational experiments for both real and artificial data;
\item Conclusions and directions for future research are given in Section~\ref{sec:conclusions}.
\end{itemize}

In Appendix we provide proofs of some technical statements and details on low and high fidelity models for the rotating disk problem.

\section{Gaussian process regression for single fidelity data}
\label{sec:gpRegression}

We consider a training sample $D = (\matX, \vecY) = \left\lbrace\vecX_i, y_i = y(\vecX_i)\right\rbrace_{i = 1}^{\sS}$,
where a point $\vecX \in \bbX \subseteq \bbR^{\iD}$ and a function value $y(\vecX) \in \bbR$. 
We assume that $y(\vecX) = f(\vecX) + \varepsilon$,
where $f(\vecX)$ is a realization of Gaussian process, and $\varepsilon$ is a Gaussian white noise with a variance $\sigma^2$. 
The goal is to construct a surrogate model for the target function $f(\vecX)$.

The Gaussian process $f(\vecX)$ is defined by its mean and covariance function
\[
k(\vecX, \vecX') = \cov (f(\vecX), f(\vecX')) = \mathbb{E} \left(f(\vecX) - \mathbb{E} (f(\vecX)) \right) \left(f(\vecX') - \mathbb{E} (f(\vecX'))\right).
\]
Without loss of generality we assume the mean value to be zero.
We also assume that the covariance function belongs to some parametric family $\{k_{\vecT}(\vecX, \vecX'), \vecT \in \Theta\subseteq \bbR^{p}\} $;
i.e., $k(\vecX, \vecX') = k_{\vecT}(\vecX, \vecX')$ for some $\vecT \in \Theta$.
Thus $y(\vecX)$ is also a Gaussian process~\cite{rasmussen2006} with zero mean and the covariance function
$\cov (y(\vecX), y(\vecX')) = k_{\vecT}(\vecX, \vecX') + \sigma^2 \delta(\vecX - \vecX')$, where $ \delta(\vecX - \vecX')$  is the Kronecker delta. 
The multivariate squared exponential covariance function~\cite{rasmussen2006} \[k_{\vecT}(\vecX, \vecX') = \theta_0^2\exp\left(-\sum_{k=1}^d\theta_k^2(x_k-x_k')^2\right)\] is widely used in applications.

The covariance function parameters $\vecT$ and the variance $\sigma^2$ fully specify the data model. We use the Maximum Likelihood Estimation (MLE) of $\vecT$ and $\sigma^2$~\cite{bishop2006pattern,rasmussen2006} to fit the model; i.e.,
we maximize the logarithm of the training sample likelihood
\begin{equation}
\label{eq:likelihood}
\log p (\vecY| \matX, \vecT, \sigma^2) = - \frac{1}{2} \left(\sS \log 2 \pi +  \log |\matK| + \vecY^T \matK^{-1} \vecY \right) \rightarrow \max_{\vecT, \sigma^2},
\end{equation}
where  $\matK = \{k_{\vecT}(\vecX_i, \vecX_j) + \sigma^2 \delta(\vecX_i - \vecX_j)\}_{i, j = 1}^{\sS}$ is the matrix of covariances between values $\vecY(\matX)$ from the training sample and $|\matK|$ is its determinant. Here
 $\sigma^2$ plays the role of a regularization parameter for the kernel matrix $ \{k_{\vecT}(\vecX_i, \vecX_j)\}_{i, j = 1}^{\sS}$, being a matrix of covariances between the values $f(\matX)$.
The recent theoretical paper~\cite{burnaev2013bernstein} and the experimental papers~\cite{bachoc2013cross,zaitsev2013properties,zaytsev2014properties} state that under general assumptions MLE parameters estimates~$\hat{\vecT}$
are accurate even if the sample size is limited and the model is misspecified.

Using estimates of $\vecT$ and $\sigma^2$ we can calculate the posterior mean and the covariances of $y(\vecX)$ at new points playing, respectively, the role of 
a~prediction and its uncertainty.
The posterior mean $\mathbb{E} (\vecY(\matX^*) | \vecY(\matX))$ at the new points $\matX^* = \{\vecX^*_i\}_{i = 1}^{\sS^*}$ has the form
\begin{equation}
\label{eq:postMean}
\hat{\vecY}(\matX^*) = \matK(\matX^*, \matX)\cdot \matK^{-1} \vecY,
\end{equation}
where $\matK(\matX^*, \matX) = \{k_{\vecT}(\vecX^*_i, \vecX_j)\}_{i = 1, \ldots, \sS^*, j = 1, \ldots, \sS}$ are the covariances between the values $\vecY(\matX^*)$ and $\vecY(\matX)$.
The posterior covariance matrix $\mathbb{V} \left(\matX^*\right) = \mathbb{E}\bigl[(\vecY(\matX^*) - \mathbb{E} \vecY(\matX^*))^T (\vecY(\matX^*) - \mathbb{E} \vecY(\matX^*))\left|\right. \vecY(\matX)\bigr]$ has the form
\begin{equation}
\label{eq:postCov}
\mathbb{V} \left(\matX^*\right) = \matK(\matX^*, \matX^*) - \matK(\matX^*, \matX)\cdot \matK^{-1}\cdot \matK(\matX, \matX^*),
\end{equation}
where $\matK(\matX^*, \matX^*) = \{k_{\vecT}(\vecX^*_i, \vecX^*_j) + \sigma^2 \delta(\vecX_i^* - \vecX_j^*)\}_{i, j = 1}^{\sS^*}$ is the matrix of covariances between the values $\vecY(\matX^*)$.

Maximum likelihood estimation of a Gaussian process regression model 
sometimes provides degenerate results --- a phenomenon closely connected to overfitting~\cite{zaitsev2013properties,zaytsev2014properties,neal1997monte,pepelyshev2010role}.
To regularize the problem and avoid inversion of large ill-conditioned matrices, one can impose a prior distribution on a Gaussian process regression model and then use Bayesian MAP (Maximum A Posteriory) estimates~\cite{burnaev2016surrogate,burnaev2015surrogate,bishop2006pattern}.
In particular in this paper we adopted the approach described in~\cite{burnaev2016surrogate}:
we impose prior distributions on all parameters of the covariance function and additional hyperprior distributions on parameters of the prior distributions. 
Experiments confirm that such approach allows to avoid ill-conditioned and degenerate cases, that can occur even more often when processing variable fidelity data.

\section{Variable fidelity Gaussian process regression}
\label{sec:variableFidelityGp}

Now we consider the case of variable fidelity data: we have a sample of low fidelity function evaluations $D_l = (\matX_l, \vecY_l) = \left\lbrace \vecX^l_i, y_l(\vecX^l_i)\right\rbrace_{i = 1}^{\sS_l}$ and a sample of high fidelity function evaluations 
$D_h = (\matX_h, \vecY_h) = \left\lbrace \vecX^h_i, y_h(\vecX^h_i) \right\rbrace_{i = 1}^{\sS_h}$
with
$\vecX^l_i, \vecX^h_i \in \bbR^{\iD}$,
$y_l(\vecX), y_h(\vecX) \in \bbR$.
The low fidelity function $y_l(\vecX)$ and the high fidelity function $y_h(\vecX)$ model the same physical phenomenon but with different fidelities.

Using samples of low and high fidelity functions values our aim is to construct a~surrogate model $\hat{y}_h(\vecX) \approx y_h(\vecX)$ of the~high fidelity function; moreover, we also would like to provide a corresponding uncertainty estimate \cite{nazarov2016,vovk2014}.

In this paper we consider a well-known variable fidelity data model (co-kriging) \cite{forrester2007multi}:
\begin{equation*}
y_l(\vecX) = f_l(\vecX) + \varepsilon_l, \,\,y_h(\vecX) = \rho y_l(\vecX) + y_d(\vecX),
\end{equation*}
where $y_d(\vecX) = f_d(\vecX) + \varepsilon_d$. 
Here $f_l(\vecX)$, $f_d(\vecX)$ are
realizations of independent Gaussian processes with zero means and covariance functions $k_l(\vecX, \vecX')$ and $k_d(\vecX, \vecX')$, respectively, and
$\varepsilon_l$, $\varepsilon_d$ are Gaussian white noise processes with variances $\sigma_l^2$ and $\sigma_d^2$, respectively.
We also set
$
\matX =
\begin{pmatrix}
\matX_l \\
\matX_h
\end{pmatrix},$
$\vecY =
\begin{pmatrix}
\vecY_l \\
\vecY_h
\end{pmatrix} \!.
$
Then the posterior mean of high-fidelity values at new points has the form
\begin{equation}
\hat{\vecY}_h(\matX^*) = \matK(\matX^*, \matX) \cdot\matK^{-1} \vecY,
\label{eq:VFGPmean}
\end{equation}
where
\begin{align*}
&\matK(\matX^*, \matX) =
\begin{pmatrix}
\rho \matK_l(\matX^*, \matX_l)\,\, &\,\, \rho^2 \matK_l(\matX^*, \matX_h) + \matK_d(\matX^*, \matX_h)
\end{pmatrix},\\
&\matK(\matX, \matX) =
\begin{pmatrix}
\matK_l(\matX_l, \matX_l) \,\,&\,\, \rho \matK_l(\matX_l, \matX_h)\\
\rho \matK_l(\matX_h, \matX_l) \,\,& \,\,\rho^2 \matK_l(\matX_h, \matX_h) + \matK_d(\matX_h, \matX_h)
\end{pmatrix},
\end{align*}
$\matK_l(\matX_a, \matX_b)$, $\matK_d(\matX_a, \matX_b)$ are matrices of pairwise covariances between $y_l(\vecX)$ and $y_d(\vecX)$ and points from some samples $\matX_a$ and $\matX_b$, respectively.
The posterior covariance matrix has the form
\begin{equation}
\mathbb{V} \left(\matX^* \right) = \rho^2 \matK_l(\matX^*, \matX^*) + \matK_d(\matX^*, \matX^*) - \matK(\matX^*, \matX)\cdot \matK^{-1}\cdot \left( \matK(\matX^*, \matX) \right)^T.
\label{eq:VFGPvariance}
\end{equation}

To estimate covariance function parameters and noise variances for the Gaussian processes $f_l(\vecX)$ and $f_d(\vecX)$ we use the following general approach~\cite{forrester2007multi}:
\begin{enumerate}
\item Estimate parameters of the covariance function $k_l(\vecX, \vecX')$ by MLE with a sample  $D = D_l$, see Section~\ref{sec:gpRegression}.
\item Calculate posterior mean values $\hat{y}_l(\vecX)$ of the Gaussian process $y_l(\vecX)$ for $\vecX\in\matX_h$,
\item Estimate parameters of the covariance function $k_d(\vecX, \vecX')$, defining the Gaussian process $y_d(\vecX)$, and $\rho$ by maximizing likelihood \eqref{eq:likelihood} with
$D = D_{\text{\rm diff}} = (\matX_h, \vecY_d = \vecY_h - \rho \hat{\vecY}_l (\matX_h))$ and $k(\vecX, \vecX') = k_d(\vecX, \vecX')$.
\end{enumerate}

As we have a big enough sample of low fidelity data, we assume that we can get accurate estimates of parameters of the covariance function $k_l(\vecX, \vecX)$, so we don't need to refine these estimates using high fidelity data. 

\section{Sparse Gaussian process regression}
\label{sec:sparse}

To perform inference for a variable fidelity Gaussian process regression we have to invert the sample covariance matrix of size $\sS \times \sS$, where $\sS = \sS_h + \sS_l$.
This operation has complexity $O(\sS^3)$, so for samples containing more than several thousands of points we cannot construct a~Gaussian process regression in a reasonable time.

In order to construct a Gaussian process regression in case of large sample sizes we propose to use an approximation to the exact inference.
The Nystr\"om approximation~\cite{burnaev11sparse,foster2009stable} of all involved positive definite matrices~$\matK(\matX^*, \matX)$, $\matK$ and $\matK(\matX^*, \matX^*)$ allows one to construct such approximation. As a basic building block of the approximation we use the Cholesky decomposition in the form similar to~\cite{foster2009stable}. This approach provides guarantees of improved numerical stability and requires reasonable amount of computations.


Let us select from the initial sample a subsample $
\matX^1 =
\begin{pmatrix}
\matX_l^1 \\
\matX_h^1
\end{pmatrix},
\vecY^1 =
\begin{pmatrix}
\vecY_l(\matX_l^1) \\
\vecY_h(\matX_h^1)
\end{pmatrix}$ of \textit{base} points with a small enough size $\sS_1 = \sS^1_h + \sS^1_l$ so we can perform the exact inference for it. The simplest, rather robust and efficient way is to perform selection without repetitions among points from the initial samples with a probability to select a point being proportional to the corresponding self-covariance value.

Hence, by definition,
\begin{align*}
&\matK_{11} =
\begin{pmatrix}
\matK_l(\matX_l^1, \matX_l^1) \,\,&\,\, \rho \matK_l(\matX_l^1, \matX_h^1) \\
\rho \matK_l(\matX_h^1, \matX_l^1) \,\,&\,\, \rho^2 \matK_l(\matX_h^1, \matX_h^1) + \matK_d(\matX_h^1, \matX_h^1)\\
\end{pmatrix}, \\
&\matK_{1} =
\begin{pmatrix}
\matK_l(\matX_l^1, \matX_l) \,\,& \,\,\rho \matK_l(\matX_l^1, \matX_h) \\
\rho \matK_l(\matX_h^1, \matX_l) \,\,&\,\, \rho^2 \matK_l(\matX_h^1, \matX_h) + \matK_d(\matX_h^1, \matX_h)
\end{pmatrix}, \\
\matK^*_{1} &=
\begin{pmatrix}
\rho \matK_l(\matX^*, \matX_l^1) \,\,& \,\,\rho^2 \matK_l(\matX^*, \matX_h^1) + \matK_d(\matX^*, \matX^1_h)
\end{pmatrix}
\end{align*}
for some new points $\matX^* = \{\vecX^*_i\}_{i = 1}^{\sS^*}$ and so we get the Nystr\"om approximation of the matrices
$\matK(\matX^*, \matX)$, $\matK$ and $\matK(\matX^*, \matX^*)$, respectively:
\begin{equation*}
\hat{\matK}(\matX^*, \matX) = \matK_1^* \matK_{11}^{-1} \matK_1, \,\,
\hat{\matK} = (\matK_1)^T \matK_{11}^{-1} \matK_1, \,\,
\hat{\matK}(\matX^*, \matX^*) = \matK_1^* \matK_{11}^{-1} (\matK_1^*)^T.
\end{equation*}

We set
\[
\matR = \begin{pmatrix}
\frac{1}{\sigma_l} \matI_{\sS_l} & 0 \\
0 & \frac{1}{\sqrt{\rho^2 \sigma_l^2 + \sigma_d^2}} \matI_{\sS_h} \\
\end{pmatrix},
\]
where $\matI_{k}$ is an identity matrix of size $k$, $\matC_1 = \matR \matK_1$ and $\matV = \matC_1 \matV_{11}^{-T}$, $\matV_{11}$ is the Cholesky decomposition of $\matK_{11}$.

\begin{theorem}
\label{th:stat1} 
For the posterior mean and the posterior covariance matrix the following Nystrom approximations hold
\begin{align}
&\hat{\vecY}^{NA}_h(\matX^*) = {\matK}^*_1 \matV_{11}^{-1} (\matI_{\sS_1} + \matV^T \matV)^{-1} \matV^T \matR \vecY, \label{eq:posteriorSparseMean}\\
&\mathbb{V}^{NA} \left(\matX^* \right) = {\matK}_1^* \matV_{11}^{-1} (\matI_{\sS_1} + \matV^T \matV)^{-1} \matV_{11}^{-T} {\matK_1^*}^T + (\rho^2 \sigma_l^2 + \sigma_d^2) \matI_{n^*}.\label{eq:posteriorSparseVar}
\end{align}
\end{theorem}

Note that there are other ways to apply the Nystr\"om approximation to the posterior covariance matrix, but these alternatives either lead to inaccurate approximations or have low numerical stability~\cite{foster2009stable}.

\begin{theorem}
\label{th:stat2}
Computational complexities of the posterior mean and the posterior covariance matrix calculations at one point using~\eqref{eq:posteriorSparseMean} and~\eqref{eq:posteriorSparseVar} are equal to $O(\sS \sS_1^2)$.
\end{theorem}
Proofs of these theorems are provided in Appendix~\ref{section:proof}.

Note that as we use the Nystr\"om approach and select base points at random from the initial sample we can get the following estimate of the approximation accuracy by directly applying results from~\cite{kumar2012sampling}.
\begin{theorem}
\label{th:nystrom_bound}
With probability $1 - \delta$ it holds that:
\begin{align*}
\frac{\|\matK(\matX^*, \matX^*) - \hat{\matK}(\matX^*, \matX^*) \|_2}{\|\matK(\matX^*, \matX^*) \|_2} &\leq 
\frac{\|\matK(\matX^*, \matX^*) - \hat{\matK}_{\sS_1}(\matX^*, \matX^*) \|_2}{\|\matK(\matX^*, \matX^*) \|_2} + \Delta, \\
\frac{\|\matK(\matX^*, \matX) - \hat{\matK}(\matX^*, \matX) \|_2 }{\|\matK(\matX^*, \matX) \|_2} &\leq 
\frac{\|\matK(\matX^*, \matX) - \hat{\matK}_{\sS_1}(\matX^*, \matX) \|_2}{\|\matK(\matX^*, \matX) \|_2} + \Delta,
\end{align*}
where $\Delta$ is of order $O\left(\frac{1}{\sqrt{\sS}}\right) O\left(\sqrt{\log \frac{1}{\delta}}\right)$, $\|\cdot\|_2$ is  $l_2$ matrix norm, 
and $\hat{\matK}_{\sS_1}(\matX^*, \matX^*)$ is the best approximation with respect to $l_2$ matrix norm, having rank~$\sS_1$.
\end{theorem}

\section{Variable Fidelity Gaussian process regression with a low fidelity function blackbox}
\label{sec:blackbox}

Suppose that we have a blackbox for the low fidelity function $y_l(\vecX)$ that estimates the low fidelity function value at any point from the design space $\bbX \subseteq \bbR^{\iD}$ on the fly. 
Let us assume that we have already constructed a Variable fidelity Gaussian process surrogate model and can calculate predictions using 
\eqref{eq:VFGPmean} and \eqref{eq:VFGPvariance}.
We can not use a huge sample of low fidelity function values due to the high computational cost of the Gaussian process regression.
Instead, in order to improve the prediction accuracy we can update the posterior mean and the posterior variance of $y_h(\vecX)$ at the new point $\vecX$ with the low fidelity function value $y_l(\vecX)$
at this point, calculated by the blackbox. 

Let us describe a computationally efficient procedure to calculate the update. We set
\[
\vecK_l(\vecX, \matX) =
\begin{pmatrix}
\matK_l(\vecX, \matX_l) \\
\rho \matK_l(\vecX, \matX_h) \\
\end{pmatrix},
\]
where $\vecX$ is the new point.
For a sample with the additional point $\vecX$ we get the expanded covariance matrix:
\[
\matK_{\text{\rm exp}} =
\begin{pmatrix}
\matK & \vecK_l \\
\vecK_l^T & k_l(\vecX, \vecX)
\end{pmatrix}.
\]

Suppose we know Cholesky factors $\matL$ and  $\matL^{-1}$ of the initial training sample covariance matrix $\matK$ and its inverse $\matK^{-1}$, respectively.
To perform computations efficiently, we update these Cholesky factors and then update the posterior mean and the posterior variance values for the expanded sample.

For the matrix $\matK_{\sS}\in\mathbb{R}^{\sS \times \sS}$ and its Cholesky decomposition using a standard approach (see Appendix~\ref{section:blackbox_chol_update}) we can get the updated Cholesky decomposition of the expanded
matrix $\matK_{\sS + 1}\in\mathbb{R}^{(\sS + 1) \times (\sS + 1)}$ in $O(\sS^2)$ steps  if the initial matrix $\matK_{\sS}$ is located in the upper left corner of the new matrix $\matK_{\sS + 1}$.
To update the inverse of the Cholesky decomposition we also need~$O(\sS^2)$ operations, as the expanded Cholesky factor is lower triangular and differs from the initial Cholesky factor only in the last row. 
Therefore, we can calculate the matrix $\matK_{\text{\rm exp}}^{-1}$ in $O(\sS^2)$ operations.

An expanded vector of covariances between the new point $\vecX$ and the initial training sample has the form
\[
\vecK_{\text{\rm exp}} =
\begin{pmatrix}
\rho \matK_l(\vecX, \matX_l) \\
\rho^2 \matK_l(\vecX, \matX_h) + \matK_d(\vecX, \matX_h) \\
\rho k_l(\vecX, \vecX)
\end{pmatrix}.
\]
We set
$
\vecY_{\text{\rm exp}} =
\left( \vecY^T, y_l(\vecX) \right)^{T},
$
where $y_l(\vecX)$ is calculated by the blackbox.
Then updated expressions for the posterior mean and the posterior variance are:
\begin{align}
&\hat{y}^{\text{\rm exp}}_h (\vecX) = \vecK_{\text{\rm exp}} \matK_{\text{\rm exp}}^{-1} \vecY_{\text{\rm exp}}, \label{eq:posteriorMeanBb}\\
&\mathbb{V}_{\text{\rm exp}} \left(\vecX \right) = \rho^2 \matK_l(\vecX, \vecX) + \matK_d(\vecX, \vecX) - \vecK_{\text{\rm exp}}^T \matK_{\text{\rm exp}}^{-1} \vecK_{\text{\rm exp}}.\label{eq:posteriorVarBb}
\end{align}
As the Cholesky factor for the updated model differs only in the last row we calculate \eqref{eq:posteriorMeanBb} and \eqref{eq:posteriorVarBb} in~$O(\sS^2)$ operations.

The total computational complexity is the sum of computational complexities of the Cholesky decomposition update and the posterior mean and variance recalculation, so for a Variable fidelity Gaussian process regression with a blackbox, representing the low fidelity function, the following theorem holds true.
\begin{theorem}
Suppose we know Cholesky factors $\matL$ and  $\matL^{-1}$ of the initial training sample covariance matrix $\matK$ and its inverse $\matK^{-1}$, respectively.
Then we can calculate the posterior mean $\hat{y}^{\text{\rm exp}}_h (\vecX)$ via \eqref{eq:posteriorMeanBb} and the variance $\mathbb{V}_{\text{\rm exp}} \left(\vecX \right)$ via \eqref{eq:posteriorVarBb} in $O(\sS^2)$ operations, where $\sS = \sS_l + \sS_h$.
\end{theorem}

As we add only one point to the initial training sample, we expect that estimates of parameters of the Gaussian process regression model do not change significantly.
While in some cases it can be reasonable to add many points, this issue raises a complex question on how and when we should re-estimate Gaussian process parameters as we add more points.
Using blackbox for the low fidelity function we can get significantly more accurate approximation with a small additional computational cost.

\section{Numerical experiments}
\label{sec:problem}

We compare four approaches for a surrogate model construction, listed below:
\begin{itemize}
\item GP --- Gaussian Process Regression using only high fidelity data;
\item VFGP --- Variable Fidelity Gaussian Process Regression using both high and low fidelity data;
\item SVFGP --- Sparse VFGP, which is a version of VFGP for the case of large training samples, introduced in Section~\ref{sec:sparse};
\item BB VFGP --- VFGP with a low fidelity function realized by a black box, introduced in Section~\ref{sec:blackbox}. 
In experiments we use the same design of experiments as in the case of VFGP, while for a surrogate model update for each new point we use a low fidelity function value at this point.
\end{itemize}

To estimate parameters of SVFGP we use a randomly selected subsample, while we use the full sample to perform approximate inference. In experiments we always use the multivariate squared exponential covariance function, see~\cite{rasmussen2006}.

We measure accuracy of surrogate models by RRMS (relative root mean square) error estimated by either the cross-validation procedure~\cite{hastie2005elements} or using a separate test sample not involved in model training process. 
RRMS error typically lies between $0$ and $1$.
RRMS error for an accurate model is close to $0$, while RRMS error for an inaccurate model is close to or greater than~$1$.
In case of a one-dimensional output and a test sample $D_{\text{\rm test}} = \{\vecX^{\text{\rm test}}_i, y_{i}^{\text{\rm test}} = f_h(\vecX^{\text{\rm test}}_i)\}_{i = 1}^{\sS_t}$ 
the RRMS error of a~surrogate model $\hat{y}(\vecX)$ is equal to
\[
RRMS(D_{\text{\rm test}}, \hat{y}) = \sqrt{\frac{\sum_{i = 1}^{\sS_t} (\hat{y}_h(\vecX^{\text{\rm test}}_i) - y^{\text{\rm test}}_i)^2}{\sum_{i = 1}^{\sS_t} (\overline{y} - y^{\text{\rm test}}_i)^2}},
\]
where $\overline{y} = \frac{1}{\sS_t} \sum_{i = 1}^{\sS_t} y^{\text{\rm test}}_i$.

In this section we assess presented approaches using several surrogate modeling problems: two artificial problems and a real problem of surrogate modeling and optimization of a rotating disk from an aircraft engine. We want to validate whether our approaches fit into requirements described in Introduction section \ref{introintro}:
we examine model construction times and accuracy of surrogate models intended to solve problems of extrapolation and interpolation.


\subsection{Toy problem}
\label{sec:artificialProblemBbVfgp}

Here we consider the well-known test problem~\cite{forrester2007multi} to construct a variable fidelity surrogate model. Data is generated by the following high fidelity $y_h(x)$ and low fidelity $y_l(x)$ functions:
\begin{align*}
y_h(x) &= (6 x - 2)^2 \sin(12 x - 4), \\
y_l(x) &= 0.5 y_h(x) + 10 (x - 1).
\end{align*}
As the problem is simple and so large samples are not required we do not perform comparison with SVFGP in this subsection. 

In order to evaluate accuracy of various algorithms we use the following procedure:
\begin{itemize}
\item Generate a high fidelity sample of size $n_h \leq 100$ with points uniformly distributed in $[0, 1]$. We consider $n_h = 6, 15$ and $30$,
\item Generate a low fidelity sample with points from the high fidelity sample and additional $(100 - n_{h})$ points uniformly distributed in $[0, 1]$,
\item Construct surrogate models using GP, VFGP and BB VFGP methods and estimate their accuracies on the test sample consisting of $1000$ high fidelity function values.
\end{itemize}
RRMS errors, provided in Table~\ref{table:rrmsArtificial}, are averaged over $50$ runs for each considered value of $n_h$.  
We see that using the low fidelity function blackbox we can significantly improve accuracy for all considered values of $n_h$.

\begin{table}
\centering
\caption{Toy problem. RRMS errors for various high fidelity training sample sizes $n_h$}
\label{table:rrmsArtificial}
\begin{tabular}{cccc}
\hline
$n_h$ & $6$ & $15$ & $30$ \\
\hline
GP       &  0.7102 & 0.0159     & $3.83e-04$ \\
VFGP     &  0.3036 & $7.42e-04$ & $1.38e-04$ \\
BB VFGP  &  0.1610 & $6.90e-07$ & $1.67e-07$ \\
\hline
\end{tabular}
\end{table}

\subsection{Artificial problem in case of large samples}
\label{sec:artificialProblemSparse}

To benchmark proposed approaches we use artificial test functions with multiple local peculiarities and input dimension $d = 6$, so we really need a rather big sample to get an accurate surrogate model. As the high fidelity function $y_h(\vecX)$ and the low fidelity function $y_l(\vecX)$ we consider
\begin{align*}
y_h(\vecX) &= 20 + \sum_{i = 1}^{d} (x_i^2  - 10 \cos (2 \pi x_i)) + \varepsilon_h,\,\vecX\in[0,1]^d, \\
y_l(\vecX) &= y_h(\vecX) + 0.2 \sum_{i = 1}^{d} (x_i + 1)^2 + \varepsilon_l,\,\vecX\in[0,1]^d.
\end{align*}
The high fidelity function is corrupted by the Gaussian white noise $\varepsilon_h$ with variance $0.001$, and
the low fidelity function is corrupted by the Gaussian white noise $\varepsilon_l$ with variance $0.002$.
We generate points in $[0,1]^d$ using Latin Hypercube Sampling~\cite{park1994optimal}.
To test extrapolation properties we limit the training sample points to the region with the range $[0, 0.5]$ instead of $[0, 1]$ for one of $6$ input variables.
The high fidelity sample size is $n_h = 100$ and the size of the subsample for SVFGP is $\sS_l^1=1000$ in all experiments.

Results are averaged over $5$ runs for each considered value of $n_l$. In Table~\ref{table:errorArtSparse} for VFGP, SVFGP, and BB VFGP approaches we provide RRMS values in the extrapolation and interpolation regimes, as well as training times.
One can see that RRMS errors of SVFGP are comparable with RRMS errors of VFGP for the same sample size, while the training time of SVFGP is tremendously smaller when the sample size is equal to $5000$, and for SVFGP the training time increases only slightly when the sample size increases.
For BB VFGP the training time in this experiment coincides with that of VFGP, while in case of $1000$ training points we get better results with BB VFGP than in case of $5000$ training points and VFGP. Also we can see that in the extrapolation regime  we get significantly better results with BB VFGP.

\begin{table}
\centering
\caption{Surrogate modeling for large samples of artificial data}
\label{table:errorArtSparse}
\begin{tabular}{clll}
\multicolumn{4}{c}{RRMS errors in case of the interpolation regime} \\
\hline
$n_l$ & 1000 & 3000 & 5000 \\
\hline
VFGP    & $0.0502 $ & $0.0170 $ & $0.0058 $ \\
SVFGP   & $0.0502 $ & $0.0305 $ & $0.0260 $ \\
BB VFGP & $0.0010 $ & $0.00029 $ & $0.00017 $ \\
\hline
\\
\multicolumn{4}{c}{RRMS errors in case of the extrapolation regime} \\ 
\hline
$n_l$ & 1000 & 3000 & 5000 \\
\hline
VFGP    & $0.3636 $ & $0.1351$ & $0.1028 $ \\
SVFGP   & $0.3636 $ & $0.3281$ & $0.3586 $ \\
BB VFGP & $0.000998$ & $0.00113 $ & $0.00034 $ \\
\hline
\\
\multicolumn{4}{c}{The training time in seconds, } \\
\multicolumn{4}{c}{Ubuntu PC, Intel-Core $i7$, $16$ Gb RAM} \\
\hline 
$n_l$ & 1000 & 3000 & 5000 \\
\hline
VFGP    & $30.46$ & $852.70$ & $7283.27$ \\
SVFGP   & $30.46$ & $33.42$ & $37.50$ \\
BB VFGP & $30.38$ & $842.97$ & $7672.60$ \\
\hline
\end{tabular}
\end{table}

\subsection{Rotating disk problem}
\label{sec:realProblem}

\subsubsection{Rotating disk model description}

\begin{figure}
        \centering
        \begin{subfigure}[b]{0.45\textwidth}
                \includegraphics[width=\textwidth]{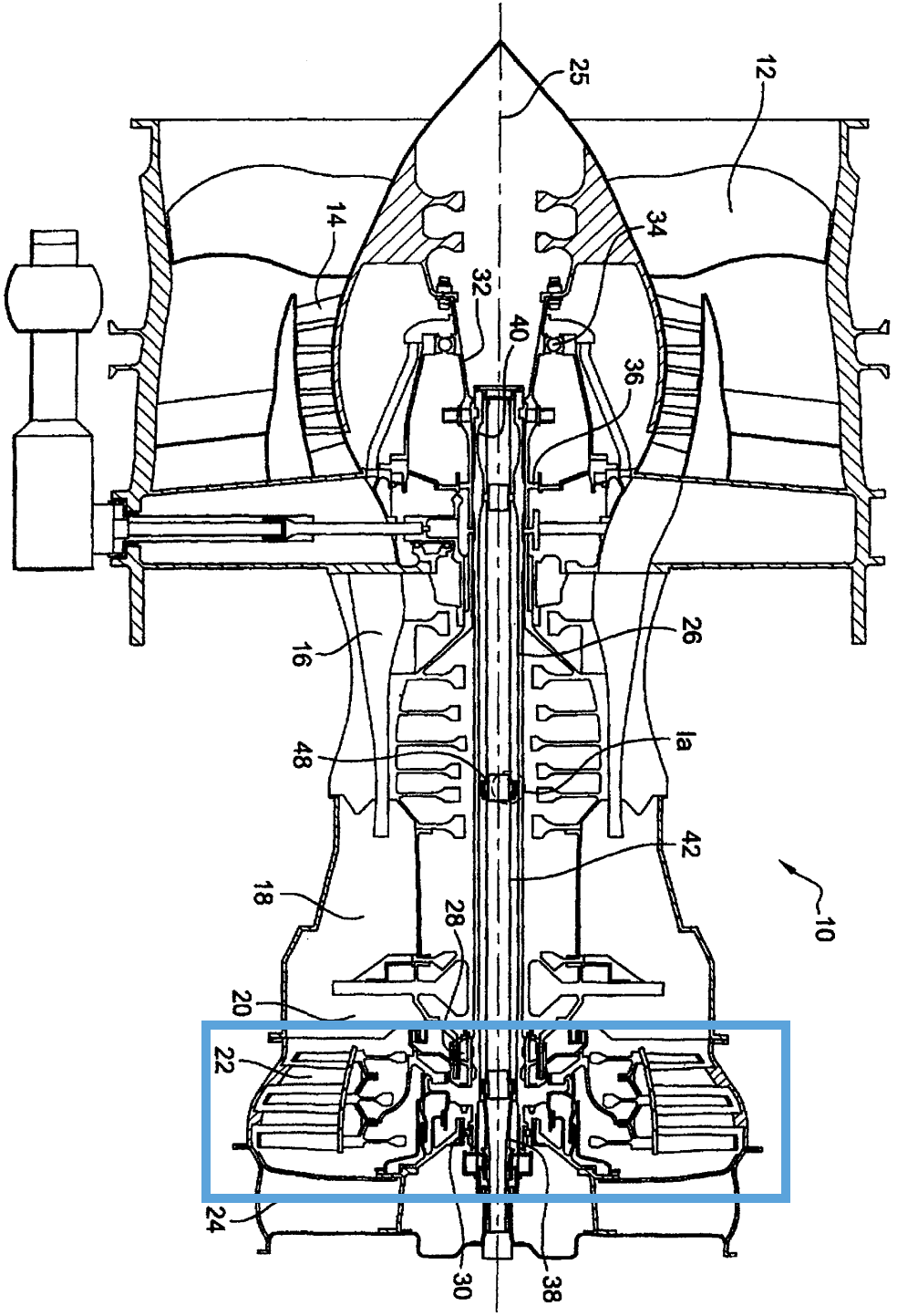}
                \caption{Aircraft engine. Rotating disk is shown by the bold rectangle at the right side of the figure}
                \label{fig:engine}
        \end{subfigure}%
        ~
        \begin{subfigure}[b]{0.40\textwidth}
                \includegraphics[width=\textwidth]{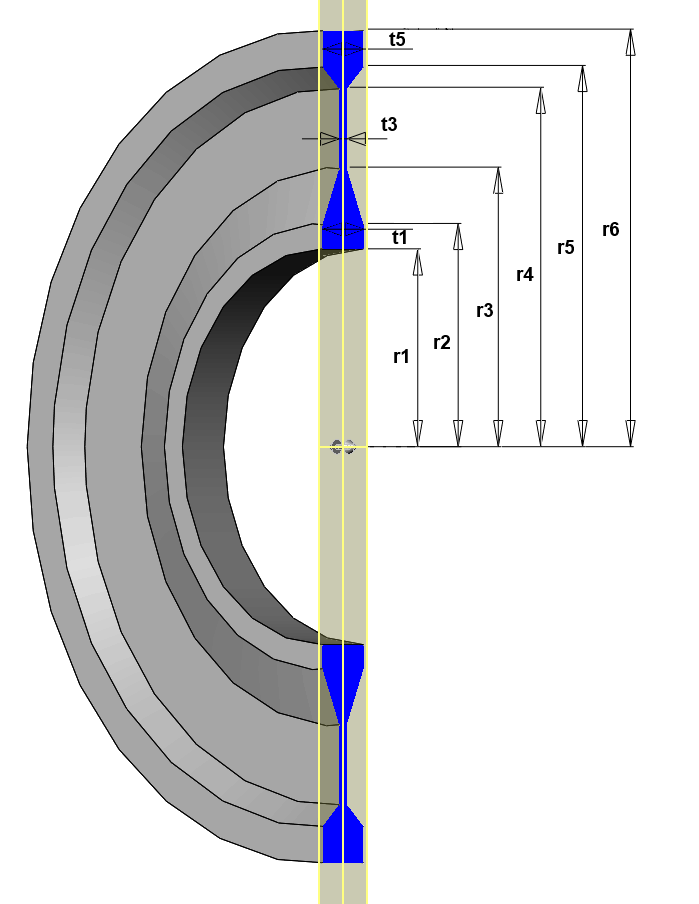}
                \caption{Rotating disk geometry}
                \label{fig:rotating_disk}
        \end{subfigure}
        \caption{Rotating disk problem}
\end{figure}

A high speed rotating risk is an important part of an aircraft engine (see Figure~\ref{fig:engine}). 
Three parameters define performance characteristics of the disk: the mass of the disk, the maximal radial displacement $u_{\text{\rm max}}$, the maximal stress $s_{\text{\rm max}}$~\cite{armand1995structural,mohan2013structural,belyaev2014exact}.
It is easy to calculate mass of the disk, as we know all geometrical parameters of the disk, while surrogate modeling of the maximal radial displacement and the maximal stress is needed since  these characteristics are computationally expensive~\cite{mohan2013structural,huang2011optimal}. 
So the focus here is on modeling the maximal radial displacement and the maximal stress.

Used parametrization of the rotating disk geometry consists of $8$ parameters: the radii $r_i$, $i = 1, \ldots, 6$, which control where the thickness of the rotating disk changes, and
the values $t_1$, $t_3$, $t_5$, which control the corresponding changes in thickness.
In the considered surrogate modeling problem we fix the radii $r_4$, $r_5$ and the thickness $t_3$ of the rotating disk, so the input dimension for the surrogate model is~$6$.
The geometry and the parametrization of the~rotating disk are shown in Figure~\ref{fig:rotating_disk}.

There are two available solvers for $u_{\text{\rm max}}$ and $s_{\text{\rm max}}$ calculation.
The low fidelity function is realized by the Ordinary Differential Equations (ODE) solver based on a simple Runge--Kutta's method.
The high fidelity function is realized by the Finite Element Model (FEM) solver.
A~single evaluation of the low fidelity function takes $\sim 0.01$ seconds, and a~single evaluation of the high fidelity function takes $\sim 300$ seconds.
More detailed comparison of the solvers is provided in Appendix~\ref{section:solvers}.

\subsubsection{Surrogate model accuracy}
\label{sec:surrogateModel}

\begin{table}
\centering
\caption{RRMS errors (with standard deviations) for the developed approaches}
\label{table:rrmsBBsampleSizeSmax}
\begin{tabular}{ccccc}
\multicolumn{5}{c}{Output $u_{\text{\rm max}}$} \\
   \hline
$n_h$   & $20$ & $40$ & $60$ & $80$ \\
   \hline
GP      & $0.287 \pm 0.039$ & $0.143 \pm 0.031$ & $0.082 \pm 0.020$ & $0.095 \pm 0.023$ \\
VFGP    & $0.212 \pm 0.075$ & $0.088 \pm 0.009$ & $0.064 \pm 0.007$ & $0.068 \pm 0.006$ \\
SVFGP   & $0.125 \pm 0.029$ & $0.074 \pm 0.016$ & $0.041 \pm 0.007$ & $0.047 \pm 0.011$ \\
BB VFGP & $0.123 \pm 0.019$ & $0.053 \pm 0.008$ & $0.030 \pm 0.007$ & $0.034 \pm 0.006$ \\
   \hline
\\
\multicolumn{5}{c}{Output $s_{\text{\rm max}}$} \\
   \hline
$n_h$   & $20$ & $40$ & $60$ & $80$ \\
   \hline
GP      & $0.505 \pm 0.10$ & $0.367 \pm 0.15$ & $0.251 \pm 0.049$ & $0.196 \pm 0.014$ \\
VFGP    & $0.363 \pm 0.07$ & $0.261 \pm 0.06$ & $0.193 \pm 0.011$ & $0.123 \pm 0.043$ \\
SFGP    & $0.190 \pm 0.06$ & $0.122 \pm 0.06$ & $0.119 \pm 0.015$ & $0.088 \pm 0.027$ \\
BB VFGP & $0.158 \pm 0.03$ & $0.162 \pm 0.03$ & $0.137 \pm 0.024$ & $0.078 \pm 0.020$ \\
   \hline
\end{tabular}
\end{table}

In this section we compare our approaches SVFGP (Sparse variable fidelity Gaussian processes) and BB VFGP (Blackbox variable fidelity Gaussian processes) with GP based only on high fidelity data and VFGP baseline methods.

We use the Latin Hypercube approach to sample points. 
The low fidelity training sample size is equal to $1000$,
the high fidelity training sample size $n_h$ is $20$, $40$, $60$, and $80$ in different experiments.
To estimate accuracy of the high fidelity function prediction we run the cross-validation procedure, applied to $140$ high fidelity data points (these points contain $n_h$ points used for training of surrogate models).  
For each fixed sample size $n_h$ we generate $5$ splits of the data into training and test samples to estimate means and standard deviations.
For SVFGP, we use $n_l=5000$ low fidelity points in total, and randomly select $n_l^1=1000$ points from them to use as base points.

The results are given in Table~\ref{table:rrmsBBsampleSizeSmax} for $u_{\text{\rm max}}$ and $s_{\text{\rm max}}$ outputs:
VFGP outperforms GP, and both SVFGP and BB VFGP outperform VFGP in terms of RRMS error.
Therefore, we should decide which one to use, SVFGP or BB VFGP, by taking into account whether the blackbox for the low fidelity function is available, or whether one uses the surrogate model in extrapolation regime, etc.

\subsection{Optimization of the rotating disk shape}

The problem is to optimize the shape of the rotating disk:
\begin{align}
\label{eq:optProblem}
m, u_{\text{\rm max}} &\rightarrow \min_{r_1, \ldots, r_6, t_1, t_3, t_5}, \\
u_{\text{\rm max}} &\leq 0.3, s_{max} \leq 600, \nonumber \\
10 &\leq r_1 \leq 110, 120 \leq r_2 \leq 140, \nonumber \\
150 &\leq r_3 \leq 168, 170 \leq r_4 \leq 200, \nonumber \\
4 &\leq t_1 \leq 50, 4 \leq t_3 \leq 50, \nonumber \\
r_5 &= 210, r_6 = 230, t_5 = 32. \nonumber
\end{align}
This problem has multiple objectives, and we are looking for a Pareto frontier, not a single solution.

Single optimization run can be described as follows:
\begin{itemize}
\item Generate an initial high fidelity sample $D_h$ with $30$ points using the Latin Hypercube sampling; 
\item Construct surrogate models using GP, VFGP, SVFGP and BB VFGP approaches using the generated high fidelity sample $D_h$ and a low fidelity sample $D_l$ of size $1000$ for GP, VFGP and BB VFGP and of size $5000$ for SVFGP; 
\item Solve multiobjective optimization problem \eqref{eq:optProblem} using the constructed surrogate models as the target functions and the constraints;
\item Using the high fidelity solver calculate true values at Pareto frontiers, constructed on the previous step, to estimate quality of the models. 
\end{itemize}
Due to properties of the used multiobjective optimization algorithm, sizes of the Pareto frontiers can slightly differ for different optimization runs, with an average size of a Pareto frontier equal approximately to $30$ points~\cite{druot2013multi}. 
So we need about $50-60$ runs of the high fidelity function to solve this optimization problem ($30$ high fidelity function evaluations to generate the initial sample and $20-30$ high fidelity function evaluations to calculate the true values at the constructed Pareto frontier). 

In order to recover a reference Pareto frontier we constructed an accurate surrogate model using $5000$ high fidelity evaluations on a uniform design over the whole design space and additional sampling in a region where points of the Pareto frontier are located.
So instead of using the high fidelity solver during optimization runs we used this surrogate model.

\begin{figure}
        \centering
        \begin{subfigure}[b]{0.45\textwidth}
                \includegraphics[width=\textwidth]{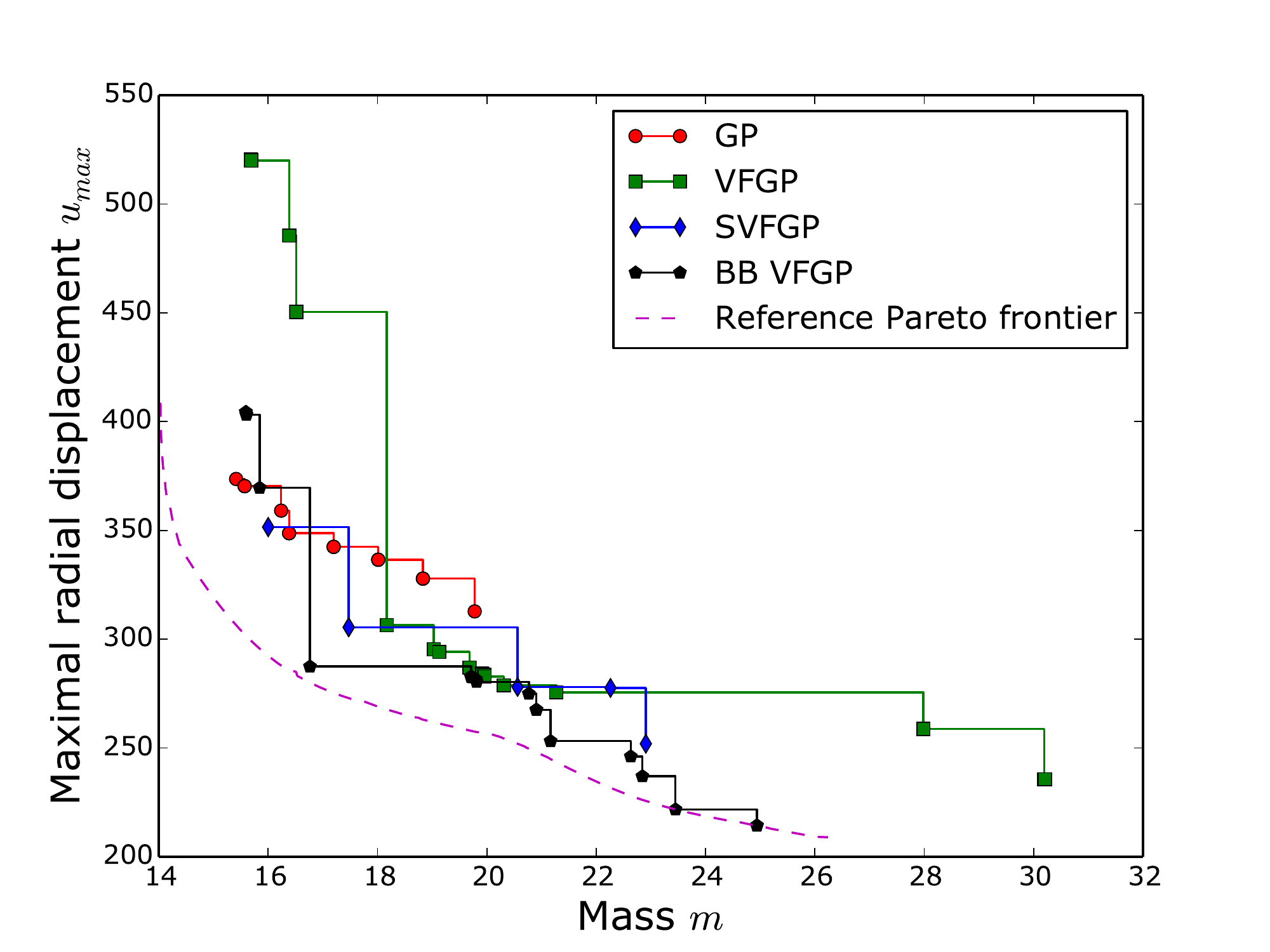}
        \end{subfigure}%
        ~
        \begin{subfigure}[b]{0.45\textwidth}
                \includegraphics[width=\textwidth]{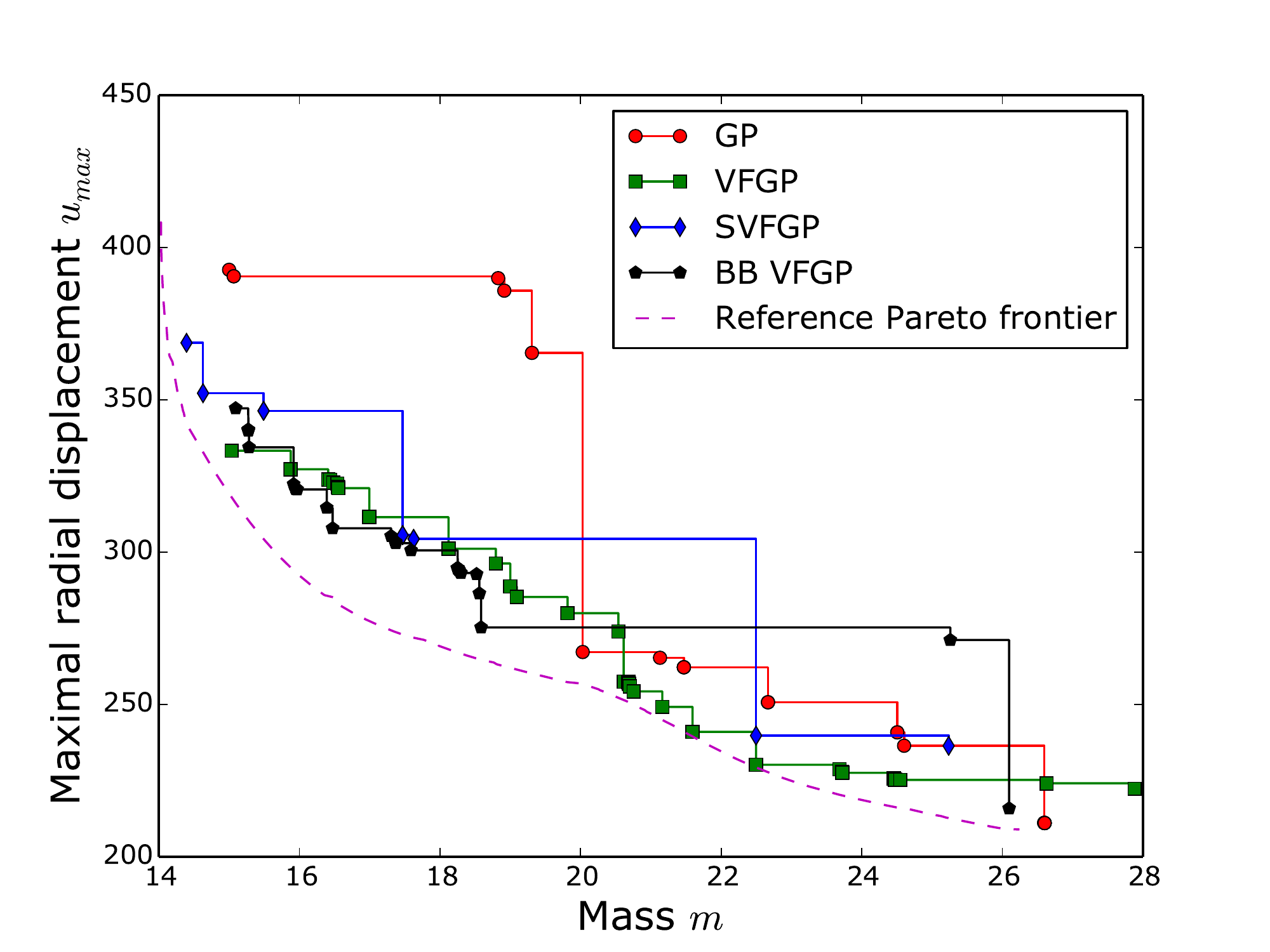}
        \end{subfigure}
        \caption{Pareto frontiers obtained by optimizing surrogate models constructed with GP, VFGP, SVFGP and BB VFGP approaches along with the reference Pareto frontier}
        \label{fig:pareto}
\end{figure}

Examples of the obtained Pareto frontiers for a single optimization run is provided in Figure~\ref{fig:pareto}.
In these runs SVFGP and BB VFGP work better than GP and VFGP.

Results of optimization are given in Table~\ref{table:optimizationResults}.
We compare minimum values of different weighted sums of the two target variables $m$ and $u_{\text{\rm max}}$ averaged over $10$ optimization runs for different initial samples.
We obtain the best value of the mass $m$ output using SVFGP algorithm and the best value of $u_{\text{\rm max}}$ using BB VFGP algorithm while optimizations based on GP and VFGP work worse.
Also, with BB VFGP we produce significantly larger amount of feasible points compared to GP, VFGP and SVFGP, which typically leads to better Pareto frontier coverage with a similar number of high fidelity solver runs. 

\begin{table}
\centering
\caption{Optimization results for different surrogate models along with minimal values for different optimization objectives. Also we indicate proportion of feasible points in a final Pareto frontier. The best values are indicated in bold font}
\label{table:optimizationResults}
\begin{tabular}{ccccc}
\hline
Objective & GP & VFGP & SVFGP & BB VFGP \\
\hline
$m$ & 16.62 & 15.69  &  \bf 15.09  & 15.63 \\
$0.8 m + 0.2 u_{\text{\rm max}}$ & 73.65 & 70.74  & 70.71 & \bf 68.10 \\
$0.6 m + 0.4 u_{\text{\rm max}}$ & 125.10 & 117.37   & 116.21  & \bf 112.55 \\
$0.4 m + 0.6 u_{\text{\rm max}}$ & 176.55 & 163.89   & 161.18  & \bf 156.99 \\
$0.2 m + 0.8 u_{\text{\rm max}}$ & 228.00 & 210.33 & 206.12 & \bf 201.44 \\
$u_{\text{\rm max}}$ & 279.44 & 256.77   & 251.05  & \bf 245.89 \\
\hline
Proportion of the feasible points & 0.54 & 0.57 & 0.55 & \bf 0.75 \\
\hline
\end{tabular}
\end{table} 

\section{Conclusions}
\label{sec:conclusions}

We presented two new approaches to variable fidelity surrogate modeling, which allow one to perform large sample inference for Variable Fidelity Gaussian process regression: 
the first approach approximates a full sample covariance matrix and its inverse; 
the second approach uses a low fidelity black box to update a surrogate model with a low fidelity function value at a point where one wants to estimate a high fidelity function, thus making usage of large low fidelity samples unnecessary.
Using developed approaches we can perform large sample inference for variable fidelity Gaussian process regression and construct more accurate surrogate models.
Our assessment of the proposed approaches by comparing them with state-of-the art methods demonstrate that we improve both accuracy of surrogate models and their training time.

Future directions of our research on surrogate modeling for variable fidelity data include 
theoretical investigation and assessing of numerical stability of the proposed approaches, their 
adaptation for the case of arbitrary number of fidelities in  data~\cite{zaytsev2016reliable}, surrogate based optimization and adaptive design of experiments.

\vspace{15px}
\appendix{{\Large \bf Appendix} }

\setcounter{section}{0}
\section{Proof of technical statements}
\label{section:proof}

In this section we provide proofs of statements from Section~\ref{sec:sparse}.
\begin{proof}[of Statement \ref{th:stat1}]
For the posterior mean we get:
\begin{align*}
\hat{\vecY}_h^{NA}(\mathbf{X}^*) &\approx \matK_1^* \matK_{11}^{-1} \matK_1^T (\matK_1 \matK_{11}^{-1} \matK_1^T + \matR^{-2})^{-1} \vecY = \matK_1^* \matK_{11}^{-1} \matK_1^T \matR (\matR \matK_1 \matK_{11}^{-1} \matK_1^T \matR + \matI_{\sS})^{-1} \matR \vecY = \\
              &= \matK_1^* \matK_{11}^{-1} \matC_1^T (\matC_1 \matK_{11}^{-1} \matC_1^T + \matI_{\sS})^{-1} \matR \vecY = \matK_1^* \matK_{11}^{-1} (\matC_1^T \matC_1 \matK_{11}^{-1} + \matI_{n_1})^{-1} \matC_1^T \matR \vecY = \\
              &= \matK_1^* (\matC_1^T \matC_1 + \matK_{11})^{-1} \matC_1^T \matR \vecY = \matK_1^* (\matC_1^T \matC_1 + \matV^T_{11} \matV_{11})^{-1} \matC_1^T \matR \vecY =\\
              &= \matK_1^* \matV^{-1}_{11} (\matV^{-T}_{11} \matC_1^T \matC_1 \matV_{11}^{-1} + \matI_{n_1})^{-1} \matV^{-T}_{11} \matC_1^T \matR \vecY = \matK_1^* \matV^{-1}_{11} (\matI_{n_1} + \matV^T \matV)^{-1} \matV^T \matR \vecY.
\end{align*}

We use the same approach to derive an equation for the posterior variance:
\begin{align*}
\mathbb{V}^{NA} \left(\mathbf{X}^* \right) - (\rho^2 \sigma_l^2 + \sigma_d^2) \matI_{n^*} &\approx \matK_1^* \matK_{11}^{-1} \matK_1^{*T} - \matK_1^* \matK_{11}^{-1} \matK_1^T (\matR^{-2} + \matK_1 \matK_{11}^{-1} \matK_1^T)^{-1} \matK_1 \matK_{11}^{-1} \matK_1^{*T}= \\
               &= \matK_1^* (\matK_{11}^{-1} - \matK_{11}^{-1} \matK_1^T (\matR^{-2} + \matK_1 \matK_{11}^{-1} \matK_1^T)^{-1} \matK_1 \matK_{11}^{-1}) \matK_1^{*T}= \\
               &= \matK_1^* (\matK_{11} + \matK_1^T \matR^2 \matK_1)^{-1} \matK_1^{*T} = \matK_1^* (\matV^T_{11} \matV_{11} + \matC_1^T \matC_1)^{-1} \matK_1^{*T} = \\
               &= \matK_1^* \matV^{-1}_{11} (\matI_{n_1} + \matV^T \matV)^{-1} \matV^{-T}_{11} \matK_1^{*T}.
\end{align*}
\end{proof}

\begin{proof}[of Statement \ref{th:stat2}]

First of all we have to calculate the matrices $\matV_{11}$ and $\matV = \matR \matK_1 \matV_{11}^{-T}$.
The matrix $\matV_{11}$ is of size $\sS_1 \times \sS_1$, so we need $O(\sS_1^3)$ to get its inverse.
To calculate $\matK_1 \matV_{11}^{-T}$ we need $O(\sS_1^2 \sS)$ operations.
Finally, as $\matR$ is a diagonal matrix, we use $O(\sS_1 \sS)$ operations to get $\matV$.

In case $n^* = 1$ to get the posterior mean we have to calculate $\matV_{11} (\matI_{\sS_1} + \matV^T \matV)^{-1} \matV^T \vecY$.
We use $O(\sS_1^2 \sS)$ operations to calculate $\matV^T \matV$,  to invert $\matI_{\sS_1} + \matV^T \matV$ we need
$O(\sS_1^3)$ operations, to calculate $\matV_{11} (\matI_{n_1} + \matV^T \matV)^{-1} \matV^T$ one uses extra $O(\sS_1^2 \sS)$ operations, and finally to calculate the posterior mean we need additional $O(\sS_1 \sS)$ operations.
Consequently, to calculate the posterior mean we use $O(\sS_1^2 \sS)$ operations.

In the same way in order to calculate $\matV_{11} (\matI_{n_1} + \matV^T \matV)^{-1} \matV_{11}^{-1}$ we need $O(\sS_1^2 \sS)$ operations to calculate $(\matI_{n_1} + \matV^T \matV)^{-1}$ and additional $O(\sS_1^3)$ operations to get the final matrix.
Consequently, in order to calculate the posterior variance we use $O(\sS_1^2 \sS)$ operations.

Finally, we need $O(\sS_1^2 \sS)$ operations to compute the required matrices, and $O(\sS_1^2 \sS)$ to obtain the posterior mean and the posterior variance from these precomputed matrices.
So, the total computational complexity is $O(\sS_1^2 \sS)$.
\end{proof}

\section{Update of the Cholesky decomposition for BB VFGP technique}
\label{section:blackbox_chol_update}

In this appendix we provide an algorithm used to update the Cholesky decomposition of the sample covariance matrix if we expand the sample with an additional point.
The problem is to evaluate the Cholesky factor $L'$ for an expanded matrix $K'$ if we know the Cholesky factor $L$ of the matrix $K$, where $K'$ and $K$ are positive definite matrices such that
  \begin{equation}
  \label{eq:Ktilde}
  K'     = \begin{pmatrix}
        K & \vecK^T \\
        \vecK & k_{(n + 1)(n + 1)} \\
        \end{pmatrix}.
  \end{equation}

We will use well-known formulas~\cite{golub2012matrix}.
Let $K\in\mathbb{R}^{n \times n}$, then $K'\in\mathbb{R}^{(n + 1) \times (n + 1)}$.
So the upper left block of the matrix $L'$ (of size $n \times n$) coincides with $L$, elements of the last row are zeros except the last element. 
Elements of the last column are:
\[
L'_{i(n + 1)} = \frac{1}{L_{ii}} \left(K'_{i(n + 1)} - \sum_{j = 1}^{i - 1} L_{j(n + 1)} L_{ji} \right), i = \overline{1, n}.
\]
A lower right element of the matrix $L'$ is 
\begin{equation}
\label{eq:lastDiagonal}
L'_{(n + 1)(n + 1)} = \sqrt{K'_{(n + 1)(n + 1)} - \sum_{j = 1}^n L_{j(n + 1)}^2}.
\end{equation}
So, we can obtain all elements of the last column by solving a system of linear equations.

It can be the case that due to numerical errors we get a negative value under the root in \eqref{eq:lastDiagonal}.
In this case a small positive value can be assigned to $L'_{(n + 1)(n + 1)}$.

Therefore if we have the inverse $L^{-1}$ of the Cholesky decomposition of $K$, 
then for the expanded matrix $K'$ the inverse of its Cholesky decomposition is:
\[
L'^{-1} = \begin{pmatrix}
    L^{-1} & \frac{-L^{-1} L'_{1:n, n + 1}}{L'_{(n + 1)(n + 1)}} \\
    0 & \frac{1}{L'_{(n + 1)(n + 1)}} \\
    \end{pmatrix},
\]
where $L'_{1:n, n + 1}$ is the last column of the matrix $L'$ without the last element.

\section{Rotating disk problem: comparison of low and high fidelity models}
\label{section:solvers}

We consider two solvers for calculation of $u_{\text{\rm max}}$ and $s_{\text{\rm max}}$.
The low fidelity function is calculated using the Ordinary Differential Equations (ODE) solver based on a simple Runge--Kutta's method.
The high fidelity function is calculated using the Finite Element Model (FEM).

\begin{figure}
        \centering
        \begin{subfigure}[b]{0.45\textwidth}
                \includegraphics[width=\textwidth]{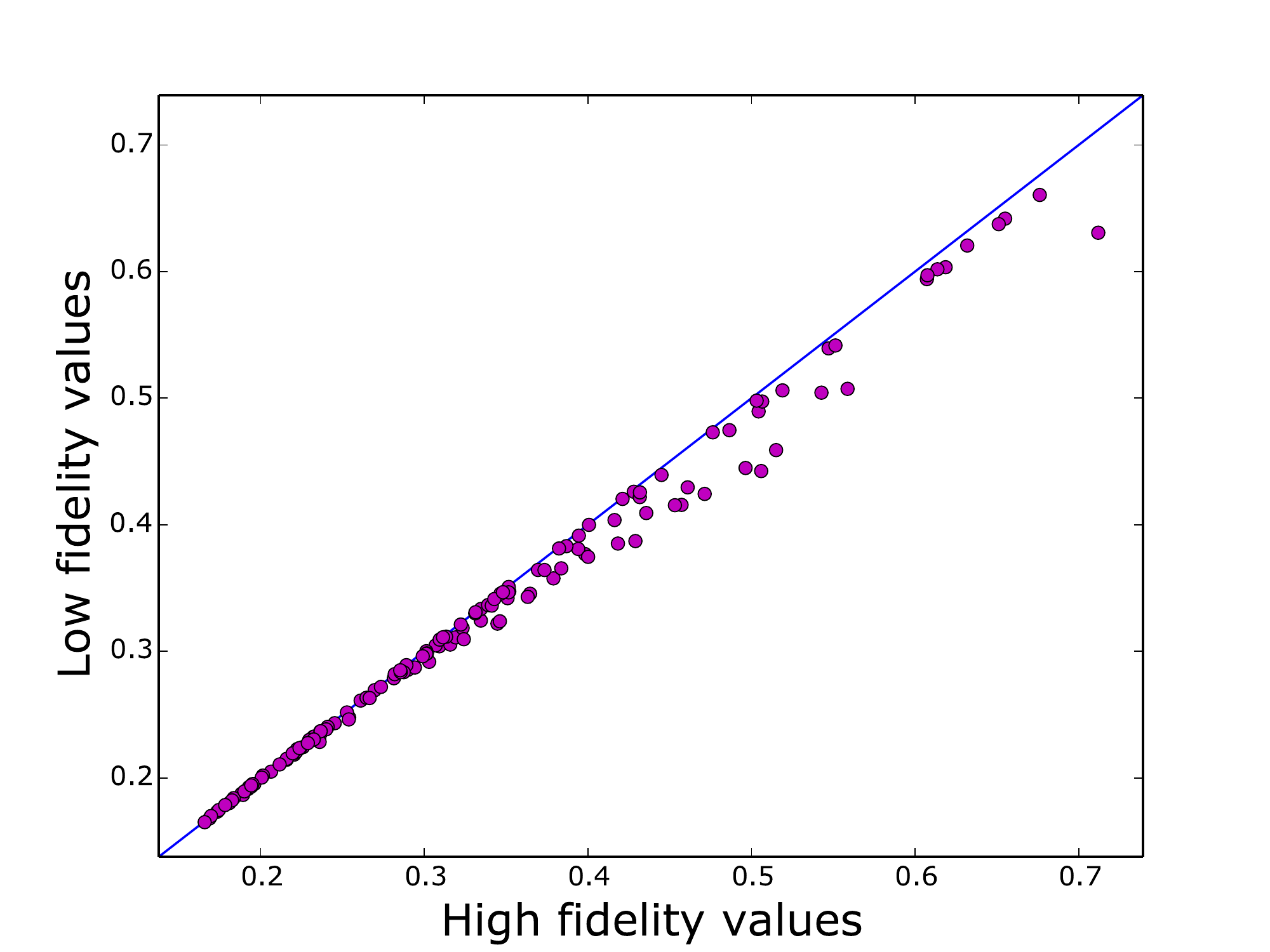}
                \caption{Scatter plot for $u_{\text{\rm max}}$}
                \label{fig:scatterVfUmax}
        \end{subfigure}%
        ~
        \begin{subfigure}[b]{0.45\textwidth}
                \includegraphics[width=\textwidth]{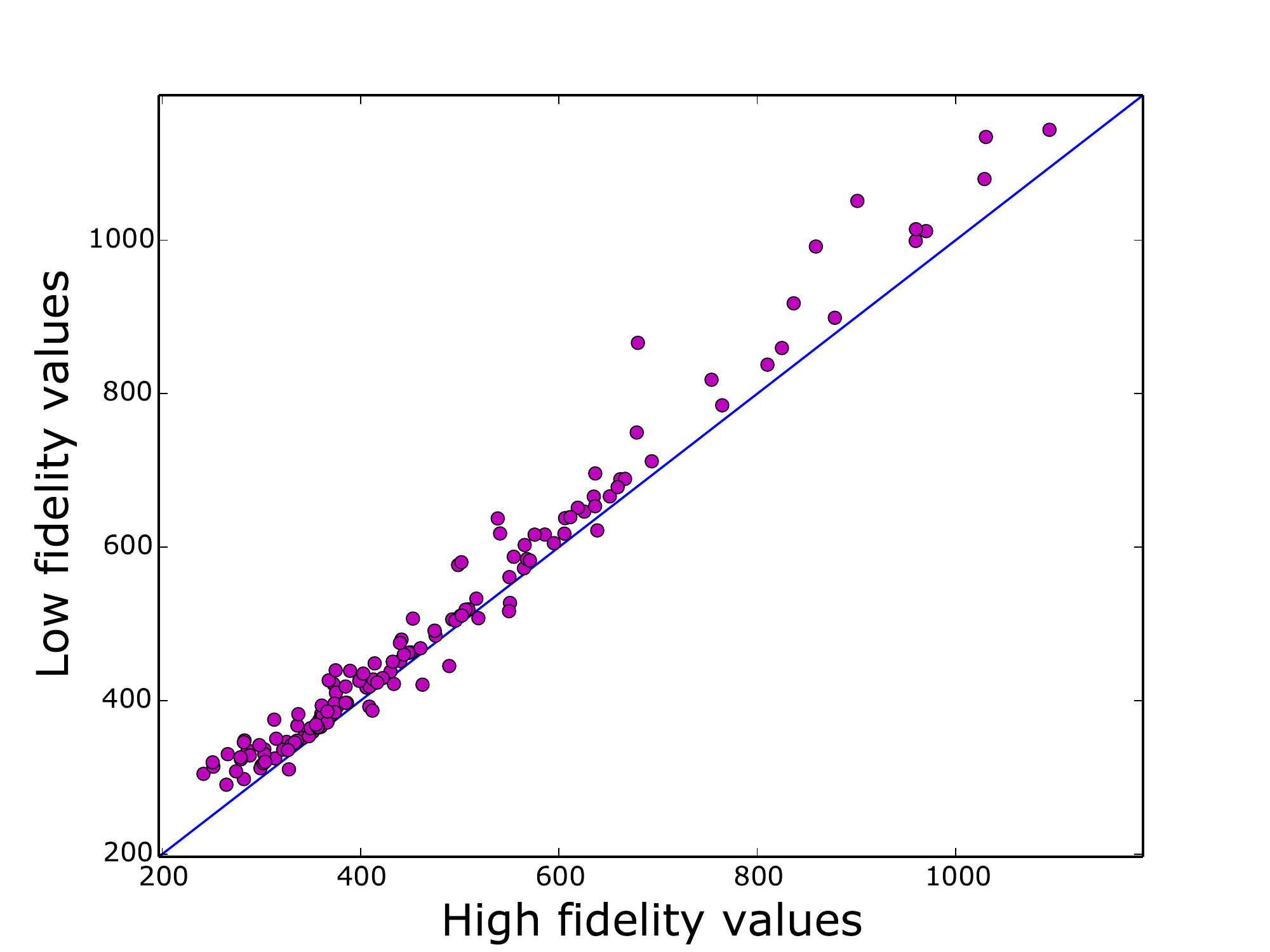}
                \caption{Scatter plot for $s_{\text{\rm max}}$}
                \label{fig:scatterVfSmax}
        \end{subfigure}
        \caption{Comparison of high and low fidelity solvers via scatter plots}
        \label{fig:scatterVf}
\end{figure}

To compare the solvers we draw scatter plots of their values and also plot slices of the corresponding functions. 
We generate a random sample of points in a specified design space box. Then we calculate low and high fidelity function values and draw the low fidelity function values versus the high fidelity function values at the same points.
The scatter plots are provided in Figure~\ref{fig:scatterVf}: the difference between values increases significantly when these values increase.

\begin{figure}
        \centering
        \begin{subfigure}[b]{0.3\textwidth}
                \includegraphics[width=\textwidth]{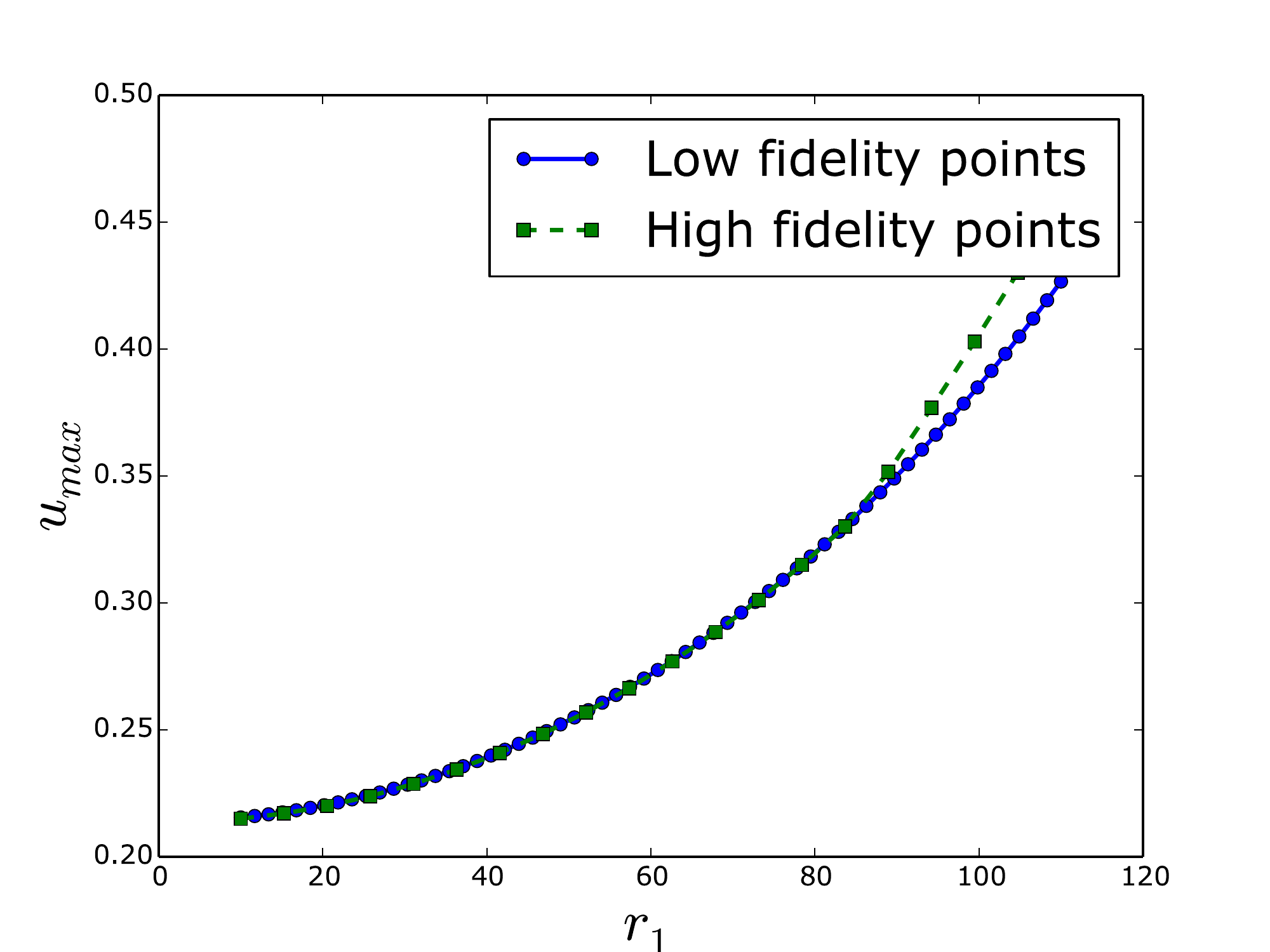}
                \caption{Slice of $u_{\text{\rm max}}$ w.r.t. $r_1$}
                \label{fig:sliceR1Umax}
        \end{subfigure}%
        ~
        \begin{subfigure}[b]{0.3\textwidth}
                \includegraphics[width=\textwidth]{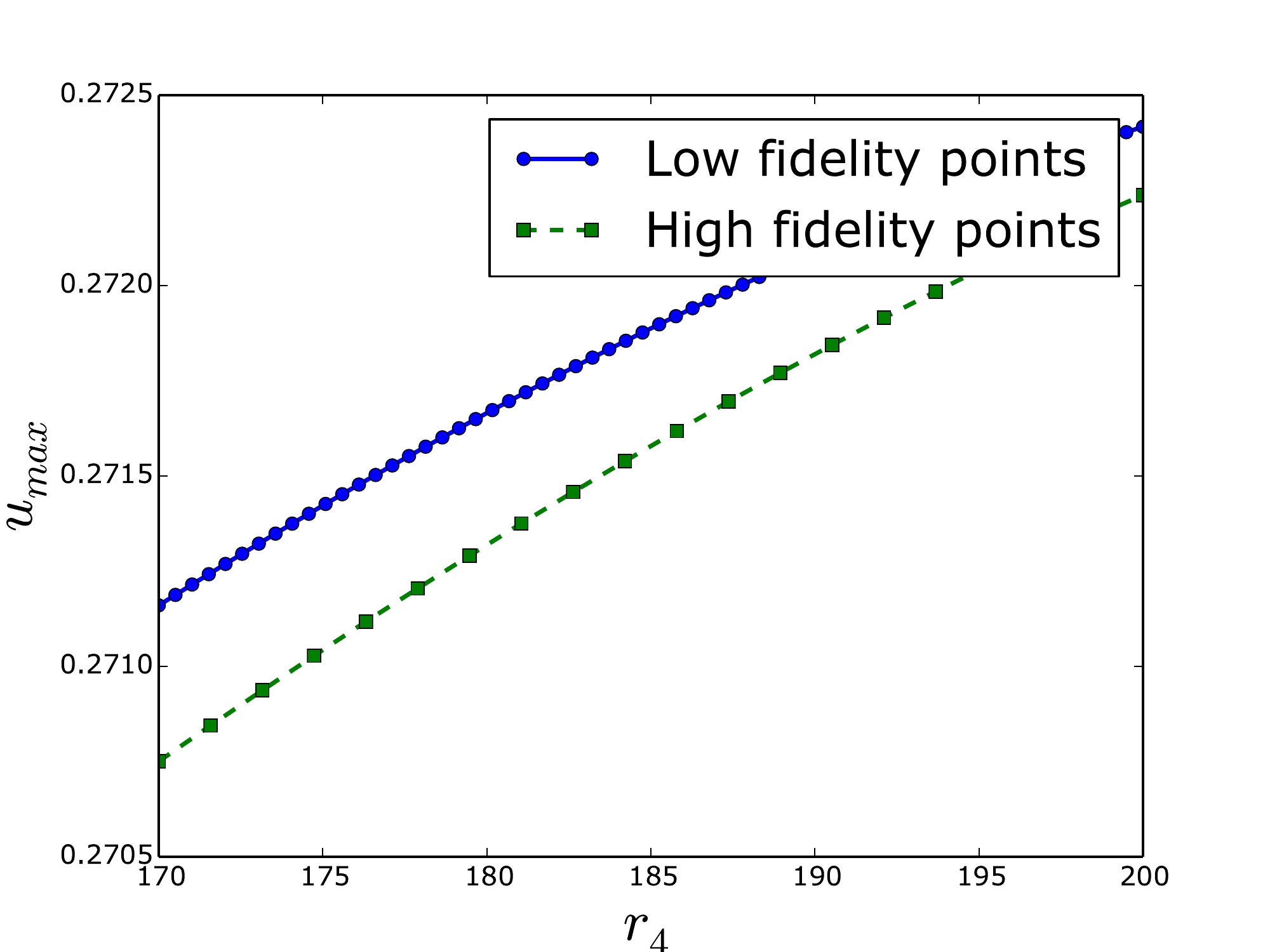}
                \caption{Slice of $u_{\text{\rm max}}$ w.r.t. $r_4$}
                \label{fig:sliceR4Umax}
        \end{subfigure}
        ~
        \begin{subfigure}[b]{0.3\textwidth}
                \includegraphics[width=\textwidth]{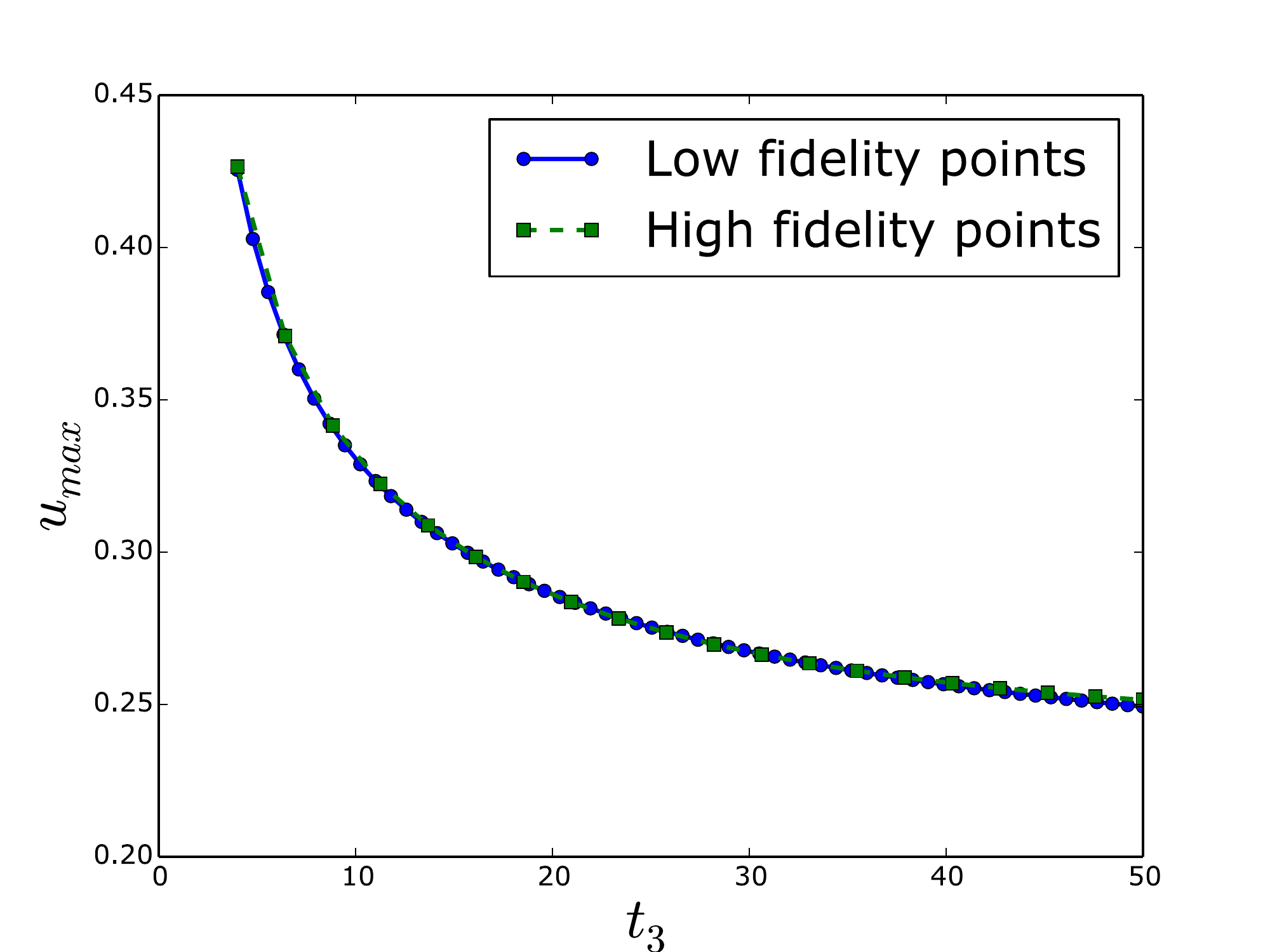}
                \caption{Slice of $u_{\text{\rm max}}$ w.r.t. $t_3$}
                \label{fig:sliceT3Umax}
        \end{subfigure}

        \begin{subfigure}[b]{0.3\textwidth}
                \includegraphics[width=\textwidth]{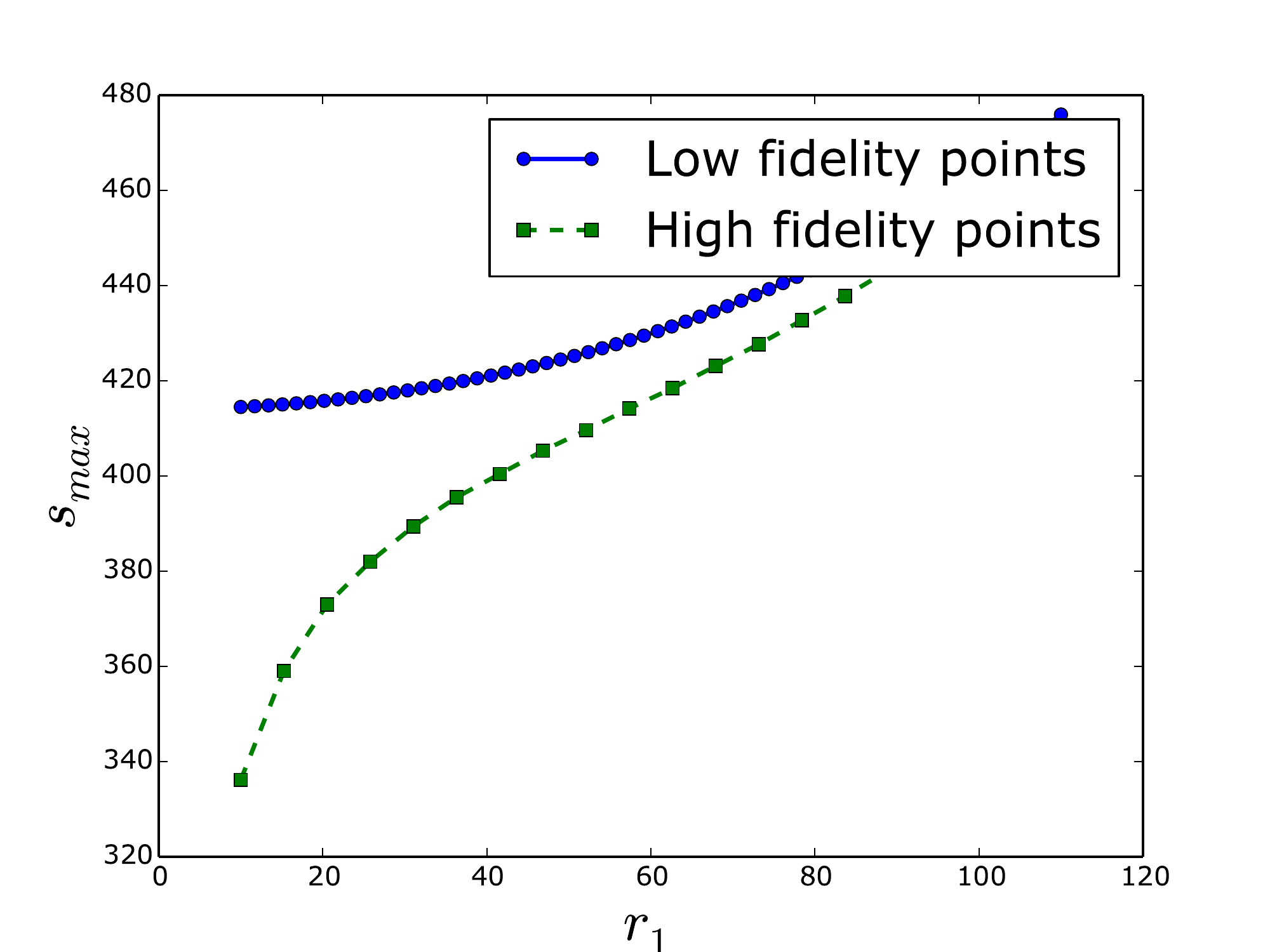}
                \caption{Slice of $s_{\text{\rm max}}$ w.r.t. $r_1$}
                \label{fig:sliceR1Smax}
        \end{subfigure}%
        ~
        \begin{subfigure}[b]{0.3\textwidth}
                \includegraphics[width=\textwidth]{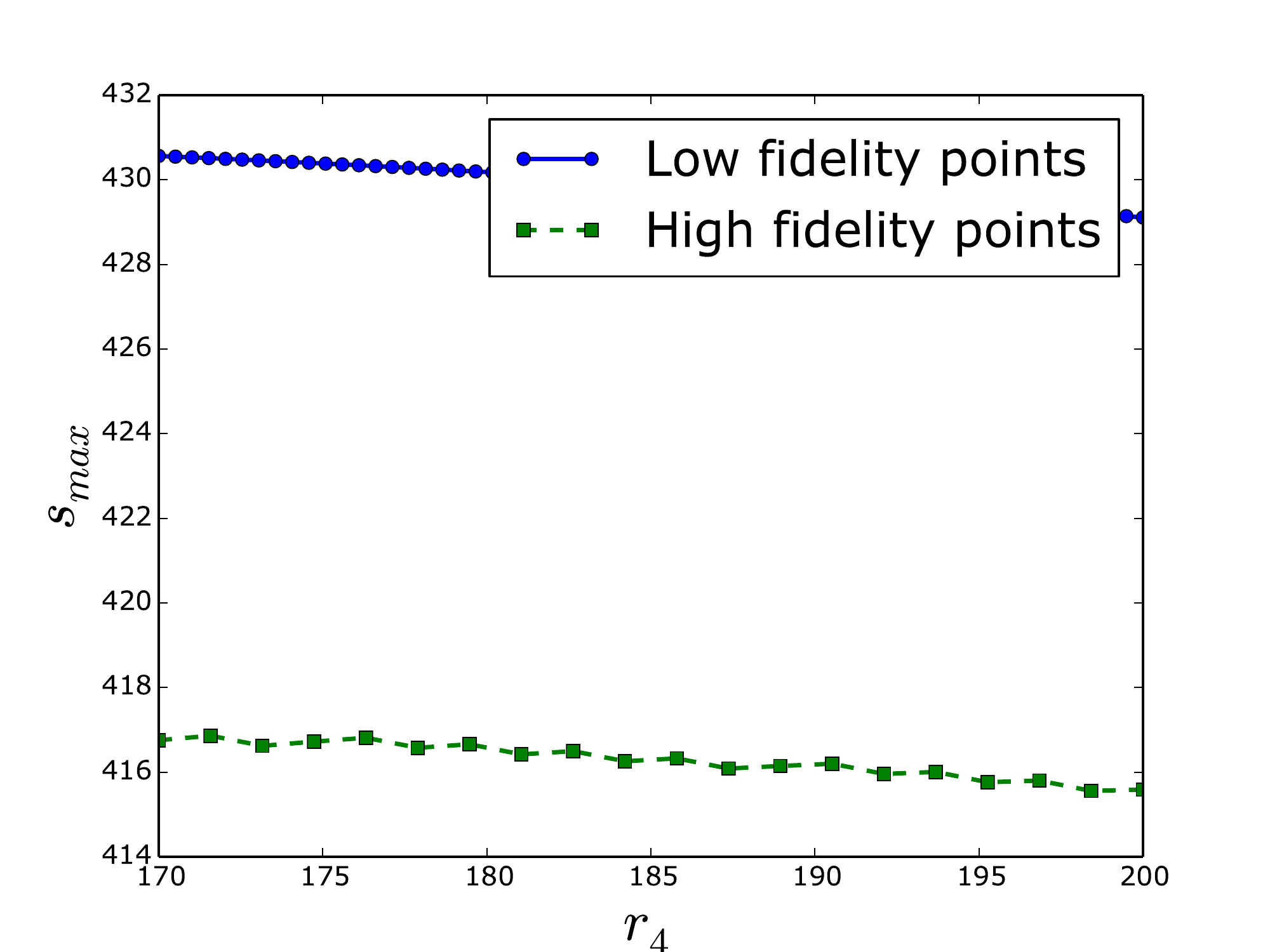}
                \caption{Slice of $s_{\text{\rm max}}$ w.r.t. $r_4$}
                \label{fig:sliceR4Smax}
        \end{subfigure}
        ~
        \begin{subfigure}[b]{0.3\textwidth}
                \includegraphics[width=\textwidth]{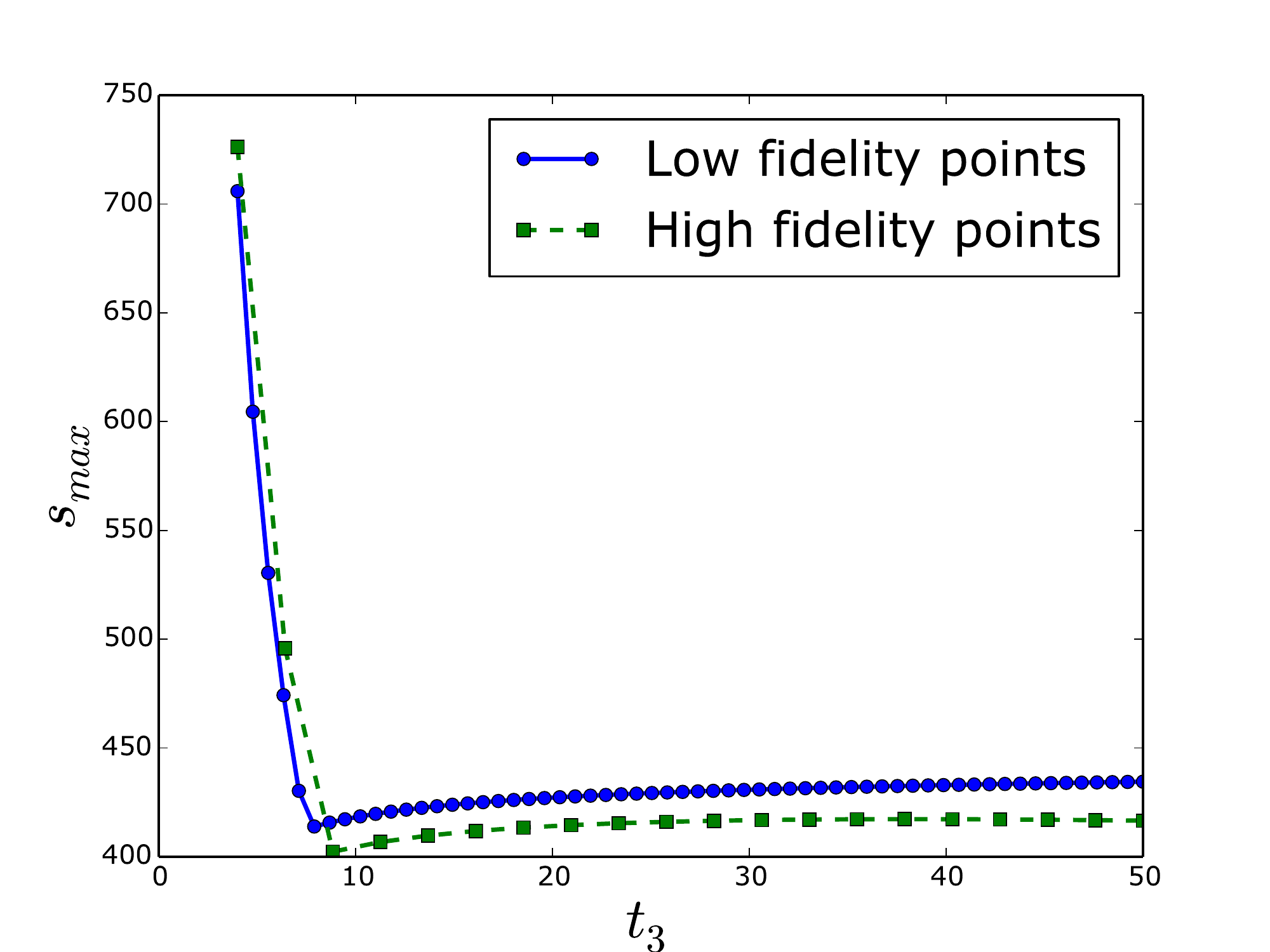}
                \caption{Slice of $s_{\text{\rm max}}$ w.r.t. $t_3$}
                \label{fig:sliceT3Smax}
        \end{subfigure}
        \caption{Comparison of high and low fidelity solvers via slices}
        \label{fig:sliceSmax}
\end{figure}

For the central point of the design space box with $r_1 = 0.06, r_2 = 0.13, r_3 = 0.16, r_4 = 0.185, t_1 = 0.027, t_3 = 0.027$ we construct one-dimensional slices by varying a single input variable in specified bounds.
Slices for different input variables for~$u_{\text{\rm max}}$ and for~$s_{\text{\rm max}}$ are given in Figure~\ref{fig:sliceSmax}.
In case of $u_{\text{\rm max}}$ the high and low fidelity functions demonstrate the same behavior, and the low fidelity function models the high fidelity function rather accurately.
For $s_{\text{\rm max}}$ the high and low fidelity functions are sometimes different: their behaviors differ for a slice along $r_1$ input variable, and local maxima differ for a slice along $t_3$ input variable.

\begin{acknowledgements}
We thank Dmitry Khominich from DATADVANCE llc for making the solvers for the rotating disk problem available. 
\end{acknowledgements}


\end{document}